\documentclass[conference]{IEEEtran}
\IEEEoverridecommandlockouts
\usepackage[letterpaper,top=0.75in,bottom=54.25bp,left=55bp,right=55bp]{geometry}

\usepackage[numbers,sort&compress]{natbib}
\usepackage{amsmath,amssymb,amsfonts}
\usepackage{graphicx}
\usepackage{capt-of}
\usepackage{placeins}
\usepackage{booktabs}
\usepackage{array}
\usepackage{multirow}
\usepackage[protrusion=false]{microtype}
\usepackage{url}
\usepackage{xcolor}
\usepackage{tikz}
\usetikzlibrary{positioning,arrows.meta,fit,backgrounds}
\usepackage[draft]{hyperref}

\newcommand{\Ep}{\ensuremath{E_{p}}}
\newcommand{\Et}{\ensuremath{E_{t}}}
\newcommand{\Eq}{\ensuremath{E_{q}}}
\newcommand{\best}[1]{\textbf{#1}}
\newcommand{\methodname}{HAT}
\newcommand{\methodmega}{Mega-HAT}
\newcommand{\methodpico}{Pico-HAT}
\newcommand{\Logv}[1]{\left[\log\!\left(#1\right)\right]^{\vee}}

\DeclareMathOperator*{\argmin}{arg\,min}
\DeclareMathOperator*{\argmax}{arg\,max}

\graphicspath{{figs/}}

\begin{document}

\title{\methodname{}: Hypothesis-Anchored Tracking for Video Monocular Spacecraft Pose Estimation}

\author{\IEEEauthorblockN{Andr\'e Lopo, Atabak Dehban, and Rodrigo Ventura}
\IEEEauthorblockA{ISR, Instituto Superior T\'ecnico, Universidade de Lisboa, Portugal\\
\texttt{\{andre.lopo,rodrigo.ventura\}@tecnico.ulisboa.pt; dehban@isr.tecnico.ulisboa.pt}}}
\hypersetup{pdfauthor={Andr\'e Lopo, Atabak Dehban, Rodrigo Ventura}}

\IEEEaftertitletext{\vspace{-24pt}
\begin{minipage}{\textwidth}
\centering
\resizebox{\textwidth}{!}{\begin{tikzpicture}[
>={Stealth[length=2.4mm]},
thumb/.style={inner sep=0pt, outer sep=0pt},
card/.style={draw=black!55, rounded corners=1.5pt, align=center,
font=\scriptsize\bfseries, inner xsep=3pt, inner ysep=2.2pt, fill=white},
band/.style={rounded corners=3pt, draw=black!25, inner sep=4pt},
bandlab/.style={font=\small\bfseries, anchor=south west},
flow/.style={->, line width=1pt, black!65},
elab/.style={font=\scriptsize, inner sep=1.5pt, fill=white, fill opacity=0.85,
text opacity=1}]

\node[thumb] (p1) at (1.75,6.55) {\includegraphics[width=24mm]{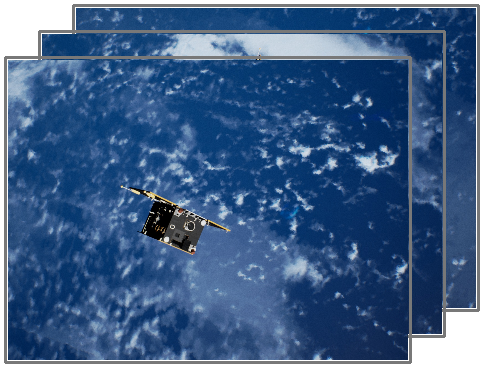}};
\node[card, fill=green!10, above=1.1mm of p1] (c1)
{Calibrated RGB sequence};
\node[thumb] (p2) at (1.75,2.95) {\includegraphics[width=32mm]{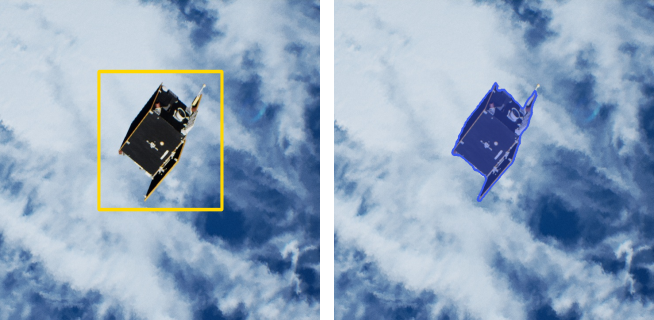}};
\node[card, fill=red!10, above=1.1mm of p2] (c2)
{\textbf{Stage 1:} detection and segmentation\\[-1pt]{\scriptsize\mdseries detection box, instance mask}};

\node[thumb] (p3) at (4.95,6.30) {\includegraphics[width=15mm]{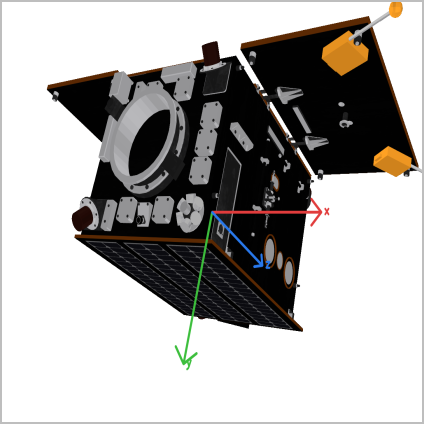}};
\node[card, fill=orange!12, above=1.1mm of p3] (c3) {CAD};
\node[thumb] (p4) at (8.15,6.30) {\includegraphics[width=36mm]{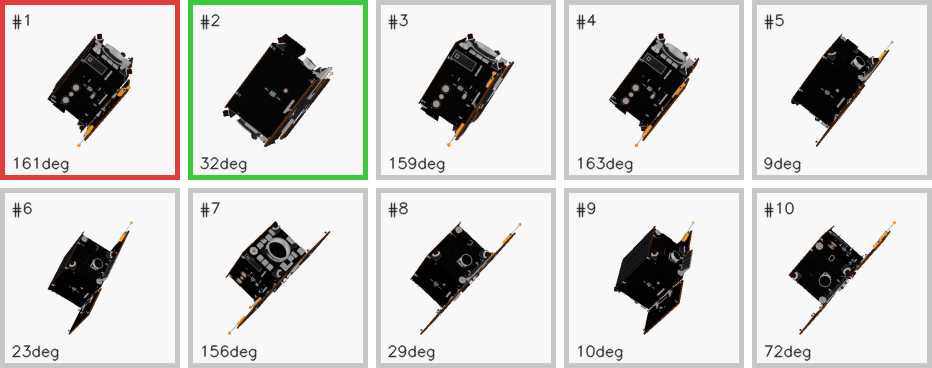}};
\node[card, fill=orange!12, above=1.1mm of p4] (c4)
{\textbf{Stage 2:} scored pose hypotheses at keyframes};

\node[thumb] (p6) at (13.6,6.25) {\includegraphics[width=46mm]{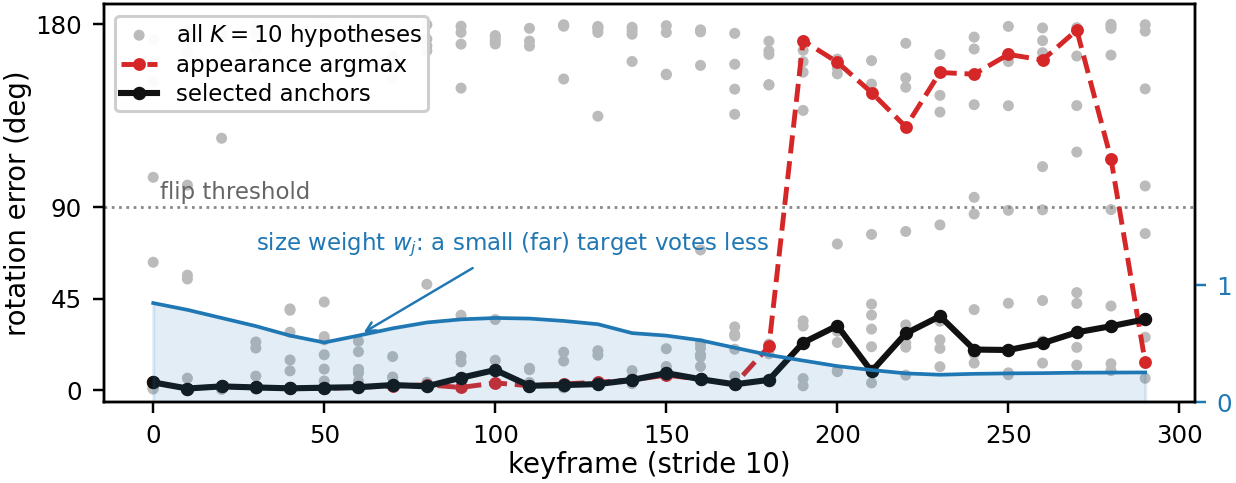}};
\node[card, fill=green!14, above=1.1mm of p6] (c6)
{\textbf{Stage 3:} causal Viterbi basin selection\\[-1pt]
{\scriptsize\mdseries appearance and motion, forward-only}};

\node[thumb] (p5) at (6.3,1.75) {\includegraphics[width=50mm]{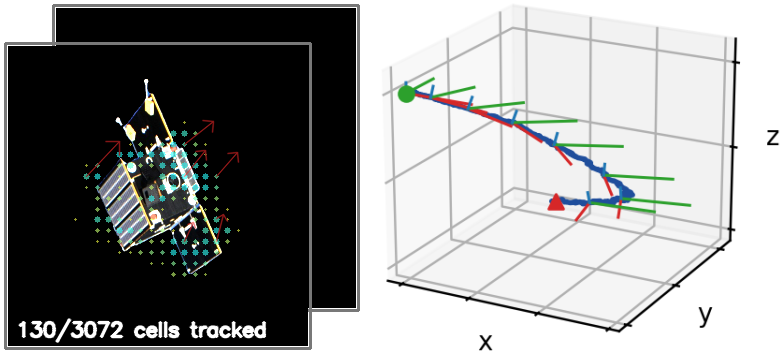}};
\node[card, fill=blue!10, above=1.1mm of p5] (c5)
{DROID-SLAM on the masked target};

\node[thumb] (p8) at (14.6,1.60) {\includegraphics[width=33mm]{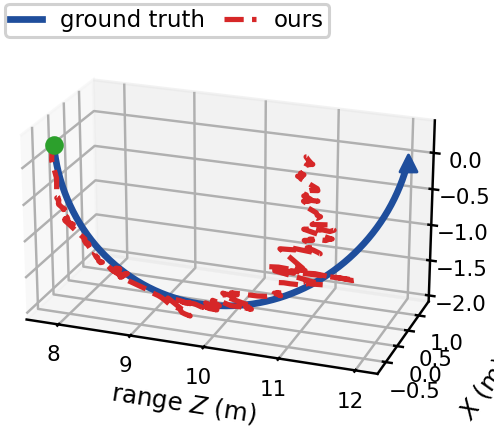}};
\node[card, fill=blue!10, above=1.1mm of p8] (c8)
{\textbf{Stage 4:} Sim(3)$+$SE(3) pose-graph fusion, then gap fill\\[-1pt]
{\scriptsize\mdseries current pose emitted, past outputs fixed}};

\begin{scope}[on background layer]
\node[band, fill=black!4,   fit=(c1)(p1)(c2)(p2)] (bA) {};
\node[band, fill=orange!7,  fit=(c3)(p3)(c4)(p4)] (bB) {};
\node[band, fill=blue!7,    fit=(c5)(p5)]         (bC) {};
\node[band, fill=green!9,   fit=(c6)(p6)]         (bD) {};
\node[band, fill=blue!7,    fit=(c8)(p8)]         (bE) {};
\end{scope}
\node[bandlab] at ([xshift=1mm]bA.north west) {(A)};
\node[bandlab, color=orange!45!black] at ([xshift=1mm]bB.north west) {(B)};
\node[bandlab, color=blue!45!black]   at ([xshift=1mm]bC.north west) {(C)};
\node[bandlab, color=green!35!black]  at ([xshift=1mm]bD.north west) {(D)};
\node[bandlab, color=blue!45!black]   at ([xshift=1mm]bE.north west) {(E)};

\draw[flow] (p1.south) -- (c2.north);
\draw[flow] (bA.east) .. controls +(1.0,0) and +(-1.0,0) .. (bB.west);
\draw[flow] (bA.east) .. controls +(1.0,0) and +(-1.0,0) .. (bC.west);
\draw[flow] (p3.east) -- (p4.west);
\draw[flow] (bB.east) -- (bD.west);
\draw[flow] (bC.north east) .. controls +(0.8,0.8) and +(-0.8,-0.8) ..
node[elab,pos=0.6]{relative rotation} (bD.south west);
\draw[flow] (bD.south) .. controls +(0,-0.7) and +(0,0.7) ..
node[elab,pos=0.55]{metric anchors} (c8.north);
\draw[flow] (bC.east) -- node[elab,above]{relative motion} (bE.west);
\end{tikzpicture}}

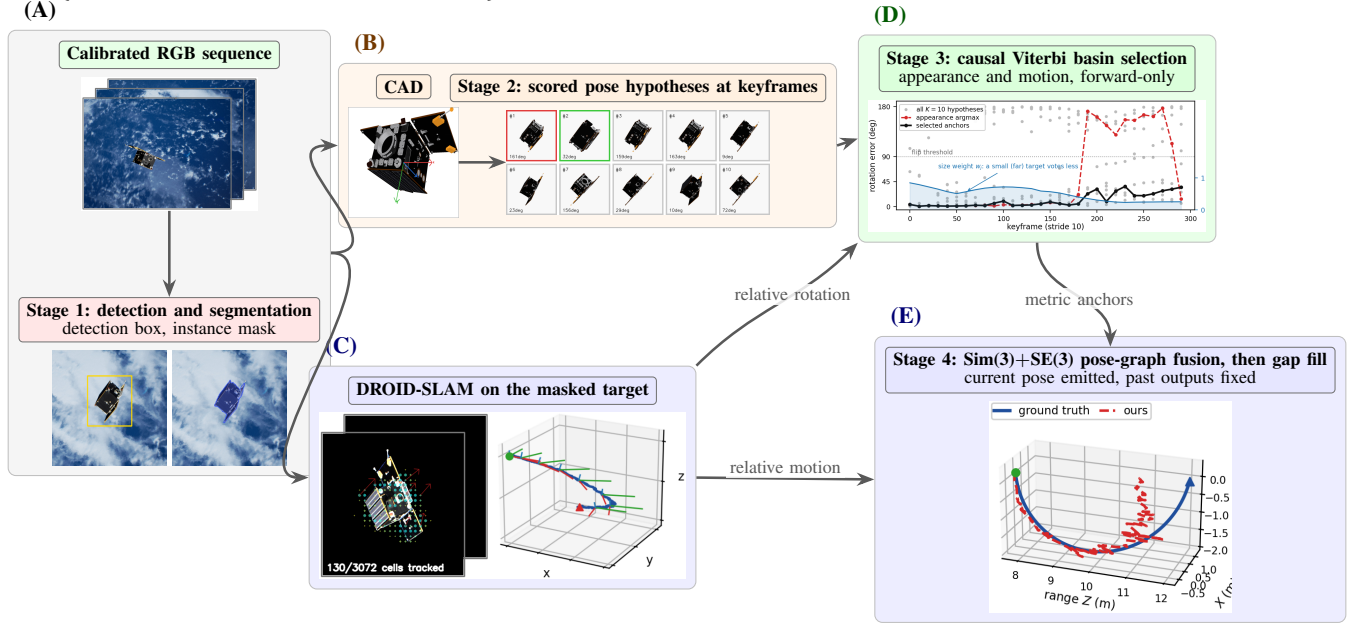
\captionof{figure}{Shared causal spacecraft-pose framework on PROBA-2 sequence RT542, using the settings in Sec.~\ref{sec:hyper}. (A) RGB detection and segmentation.
(B) CAD-based hypotheses at scheduled keyframes. In this example, appearance rank~1 is a
$161^\circ$ near-symmetric error. The selected pose is rank~2 ($32^\circ$).
(C) Masked-target DROID estimates relative motion. (D) Size-weighted forward
selection retains a consistent orientation basin. (E) Fusion and causal DROID produce a pose at every frame.}

\label{fig:pipeline}
\end{minipage}
\par\vspace{8pt}
}

\maketitle

\setlength{\abovedisplayskip}{3pt plus 2pt minus 1pt}
\setlength{\belowdisplayskip}{3pt plus 2pt minus 1pt}
\setlength{\abovedisplayshortskip}{0pt plus 2pt}
\setlength{\belowdisplayshortskip}{3pt plus 2pt minus 1pt}
\setlength{\jot}{2pt}

\begin{abstract}
Monocular 6-DoF pose estimation of non-cooperative targets is important for
on-orbit servicing and debris removal. A single-image estimator can confuse
near-symmetric spacecraft orientations, and tracking can preserve an incorrect pose.
We present Hypothesis-Anchored Tracking (\methodname{}), a causal framework
that uses inter-frame motion to select among competing CAD-based pose
hypotheses before alignment and fusion. Rather than independently choosing
the highest-scoring hypothesis in each image, \methodname{} retains competing
orientation histories and selects a pose to anchor the relative trajectory
estimated by monocular SLAM. Sparse anchors and pose fusion provide per-frame
estimates after initialization without revising past outputs. The method requires
only a calibrated RGB sequence, a metric CAD model, and target image regions,
which can be supplied by detection or segmentation. The pretrained pose and
SLAM networks require no target-specific training or fine-tuning.
We evaluate two versions, \methodmega{} and \methodpico{}, using MegaPose and PicoPose,
on SPARK-2024, SwissCube and SHIRT, with YCB-Video assessing
performance outside the space domain. Using one temporal configuration per
method, the arithmetic means of the four dataset-wise comparisons show
$9.4\%$ lower mean pose error and $3.76\times$ the sustained input $\mathrm{FPS}$ for
\methodmega{} relative to independent MegaPose, and $23.9\%$ lower mean pose
error and $2.42\times$ the $\mathrm{FPS}$ for \methodpico{} relative to independent
PicoPose.
Mega-HAT ablations on SPARK and an offline reference examine component
contributions and the effect of revising past estimates.

\end{abstract}


\section{Introduction}
On-orbit servicing and active debris removal require the pose of a target
spacecraft relative to an approaching chaser. We consider a non-cooperative
target observed by a calibrated monocular camera, without fiducials or
telemetry. A single camera reduces sensing mass, power and system complexity
compared with stereo or active ranging~\cite{pauly2023survey}.

Spacecraft pose estimation is hard under harsh illumination, with deep shadows
and bright reflections obscuring surface details. At long range, the target
covers few pixels, making these details harder to distinguish. Near-symmetric
structures can then look similar at very different orientations. A single-image
estimator may select a plausible but incorrect orientation. Independent
frame-wise decisions discard motion evidence, while local tracking can
continue following the incorrect orientation.

Our framework, \methodname{}, retains competing orientation histories and
uses inter-frame motion to select an absolute pose before alignment and fusion.
The selected poses anchor a relative trajectory estimated by monocular SLAM.
A metric CAD model supplies the scale that monocular imagery alone cannot
determine. We implement two versions using MegaPose~\cite{labbe2022megapose}
or PicoPose~\cite{picopose2025} as the absolute-pose estimator, named
\methodmega{} and \methodpico{}, respectively. Both use the same temporal
framework, with separate parameter settings and no target-specific training
or fine-tuning of the pretrained networks.

This work presents three main contributions:
\begin{itemize}
\item \textbf{Causal multi-hypothesis selection before fusion.}
Our main contribution is to retain competing orientation histories and use
relative motion to select the current absolute-pose anchor before alignment
and fusion. The decoder retains a score for each current candidate rather
than committing to a single orientation history. Fusion then combines the
selected anchor with relative motion. This
ordering addresses incorrect orientations that local tracking alone can preserve.
\item \textbf{Evaluation across datasets and absolute estimators.}
We evaluate both versions on three spacecraft benchmarks and
YCB-Video~\cite{xiang2018posecnn}, using one temporal configuration per version across datasets.
\item \textbf{Component ablations and online/offline controls.}
We use component ablations on \methodmega{} and SPARK to identify the contribution
of selection before fusion and the smaller, configuration-dependent effect
of size weighting. Online/offline controls examine the effect of future
observations and retained state.
\end{itemize}

\section{Related Work}
\label{sec:related}
\textbf{Monocular spacecraft pose estimation.} The dominant approach regresses
predefined 3D landmarks and solves PnP against known geometry~\cite{chen2019spnet}. Such
methods have achieved leading results on the SPEED, SPEED+, and SPARK-2024
benchmarks~\cite{kisantal2020speed,park2022speedplus}, but require landmarks and training
images from the target spacecraft and remain sensitive to
the synthetic-to-real gap~\cite{liu2024domaingap,park2024shirt}.

\textbf{Temporal and sequence-based methods.} Prior work uses navigation
filters~\cite{park2023ukf}, sequence networks~\cite{rondao2022chinet}, and cross-domain
tracking~\cite{liu2024domaingap}. \citet{sosa2025motion} add optical flow
to a target-trained ViT to localize spacecraft keypoints. \citet{zhang2024selfiter}
iteratively refine a reconstructed keypoint model and a multi-task prediction network,
then impose motion consistency.
Our method explicitly retains competing orientation hypotheses before fusion,
and Sec.~\ref{sec:filters} compares discrete selection with continuous-state filters.

\textbf{Novel-object pose.} MegaPose~\cite{labbe2022megapose} uses
render-and-compare networks for coarse pose ranking and iterative refinement
of unseen objects. PicoPose~\cite{picopose2025} instead establishes
pixel correspondences with rendered templates, aligns them through a global
affine estimate, and refines local correspondence offsets before
PnP/RANSAC. We use its public implementation and pretrained models in \methodpico{}.
GigaPose~\cite{nguyen2024gigapose} retrieves template orientations and
estimates the remaining pose parameters from patch correspondences.
GenFlow uses render-to-image flow~\cite{moon2024genflow}.
The closest spacecraft system,
CroSpace6D~\cite{zuo2024crospace6d}, combines Mask2Former, MegaPose,
ORB-SLAM3 and similarity alignment, but reconstructs a target mesh from
training images. In contrast, our framework uses \emph{causal selection}
among competing absolute orientations before similarity alignment and
pose fusion.

\textbf{RGB video tracking.} SRT3D~\cite{stoiber2022srt3d} estimates pose
from sparse contour correspondence lines and foreground/background appearance.
It provides a classical region-based control with an automatic pose initializer.
RGBTrack~\cite{guo2025rgbtrack} builds on FoundationPose, estimates initial depth
from a metric CAD model and mask, and combines XMem with Kalman-based recovery.
We evaluate its public RGB-only initialization and recovery paths.

\textbf{Visual SLAM.} DROID-SLAM~\cite{teed2021droid}
uses dense correspondences and a differentiable factor graph. We combine
its motion estimates with discrete $\mathrm{SO}(3)$ hypotheses through Viterbi
decoding~\cite{viterbi1967}.

\section{Methodology}

\subsection{Framework Overview}
We estimate the pose of a rigid target spacecraft in the observing camera
frame. The inputs are calibrated RGB images, a metric CAD model, and target
boxes or masks obtained without future observations. Stage~1 supplies these image
regions to the absolute and relative branches (Fig.~\ref{fig:pipeline}).
The pose and SLAM networks use pretrained
weights without target-specific training or fine-tuning.
The absolute branch proposes metric target-in-camera poses at scheduled
keyframes; SLAM estimates relative motion in its own coordinate frame and
scale. Stage~3 uses appearance and motion to select a pose, called an anchor.
Stage~4 uses these anchors to align SLAM with metric target coordinates and
combines both estimates. At frame $t$, every stage uses only frames $\le t$,
and no pose already returned as an output for a processed frame is revised.
\subsection{Stage 2: Keyframe Pose Hypothesis Generation}
\label{sec:coarse}
At keyframe $j$, a single-image pose estimator returns up to $K$ hypotheses
$\{(M_{j,k},\ell_{j,k})\}_{k=1}^{K}$: metric poses $M_{j,k}\in\mathrm{SE}(3)$
in a common object frame and max-centered appearance scores $\ell_{j,k}$. The estimator
may use render-and-compare inference, correspondences, or another
CAD-based procedure. Any internal refinement precedes temporal selection.
Keyframes follow a fixed cadence or minimum spacing among available
detections, with invalid candidates excluded. Stage~4 supplies inter-anchor
poses.

\subsection{Target-Relative Motion Estimation}
\label{sec:slam}
We run DROID-SLAM~\cite{teed2021droid} incrementally with local optimization,
without global trajectory optimization, and keep each per-frame estimate fixed
once emitted. SPARK and SwissCube use entire frames with
the background suppressed. YCB-Video uses unmasked images because the evaluated
objects and background form a largely static scene viewed by a moving camera.
SHIRT's predominantly black background needs no additional masking.
When tracking a rigid target or a scene containing a stationary target,
the states $T^{S}_i=(R^{S}_i,t^{S}_i)$ place the camera in a target-fixed
SLAM frame, opposite to the absolute-pose convention. Stage~4 therefore inverts
them. Both implementations use the rotation increment defined in
Eq.~\eqref{eq:increment} for selection.
Absolute SLAM orientation is not used directly because its gauge is arbitrary and it
can drift. Background suppression encourages the relative branch to
follow the target. Its effect is evaluated in Sec.~\ref{sec:maskablation}.

\subsection{Stage 3: Causal Orientation Basin Selection}
\label{sec:resolver}
Near-symmetric spacecraft admit competing orientation basins, or groups of
nearby orientations. The decoder evaluates candidate orientation histories using
appearance scores and inter-keyframe motion consistency. For each current
candidate, it retains the highest cumulative score among all histories whose
final pose is that candidate. The candidate with the highest cumulative score
is proposed as the current anchor. This can recover a mis-ranked pose, but not
one absent from the candidate set.

Let $j$ index scheduled keyframes, $i_j$ their frame indices, and $k$ the
available candidates. Pose $M_{j,k}\in\mathrm{SE}(3)$ maps target to camera coordinates,
with rotation $R_{j,k}$ and appearance score $\ell_{j,k}$, supplied by the
single-image estimator (higher is better). The methods
use the scaled appearance score
\begin{equation}
\label{eq:appearance}
\tilde{\ell}_{j,k}=\alpha\left(\ell_{j,k}-\max_h\ell_{j,h}\right),
\end{equation}
where $\alpha>0$ is a scaling factor chosen for each method to bring MegaPose
and PicoPose appearance scores to similar ranges. Both configurations use
equal size weights; Sec.~\ref{sec:ablation} defines the size-dependent
extension.

SLAM rotations map camera coordinates into a target-fixed frame. The increment
\begin{equation}
\label{eq:increment}
\Delta_j^S=(R_{i_{j+1}}^S)^\top R_{i_j}^S
\end{equation}
left-multiplies $R_{j,k}$ to predict the next orientation. With $d$ the smallest
rotation angle in degrees and $\mu>0$ the motion weight, a candidate path
$\pi=(\pi_0,\ldots,\pi_J)$ through observed keyframes maximizes
\begin{equation}
\label{eq:viterbi}
\sum_{j=0}^{J}\tilde{\ell}_{j,\pi_j}
-\mu\sum_{j=0}^{J-1}
d\!\left(R_{j+1,\pi_{j+1}},\Delta_j^S R_{j,\pi_j}\right).
\end{equation}
The forward Viterbi recurrence~\cite{viterbi1967} is
\begin{equation}
\label{eq:forward}
\begin{aligned}
V_j(k) &= \tilde{\ell}_{j,k}+\max_h\bigl[V_{j-1}(h)\\
&\qquad{}-\mu d\!\bigl(R_{j,k},\Delta_{j-1}^S R_{j-1,h}\bigr)\bigr],
\end{aligned}
\end{equation}
with $V_0(k)=\tilde{\ell}_{0,k}$ and no motion penalty for a missing increment.
The proposal is $k_j=\argmax_k V_j(k)$. Selecting this anchor does not discard
the other endpoint scores: they can support a different orientation history
at a later keyframe.

\subsection{Stage 4: Pose Fusion and Gap Handling}
\label{sec:fusion}
Stage~4 first aligns SLAM to metric target coordinates, then estimates the
current pose from absolute and relative measurements. Let $i$ index fusion
updates, $M_i=(R_{M_i},t_{M_i})$ denote an available target-in-camera anchor,
and $T_i^S=(R_i^S,t_i^S)$ place the camera in the SLAM frame.

\textbf{Metric alignment.} The similarity $A=(s,R^A,t^A)$ converts SLAM
distances to meters with $s>0$, rotates SLAM axes into target axes with
$R^A\in\mathrm{SO}(3)$, and locates the SLAM origin at $t^A\in\mathbb R^3$.
The aligned camera pose $(R_i^F,t_i^F)$ and its target-in-camera inverse
$S_i(A)=(R_{S_i},t_{S_i})$ are
\begin{equation}
\label{eq:alignmentmap}
\begin{alignedat}{2}
R_i^F &= R^A R_i^S, &\qquad t_i^F &= sR^A t_i^S+t^A,\\
R_{S_i} &= (R_i^F)^\top, & t_{S_i} &= -(R_i^F)^\top t_i^F.
\end{alignedat}
\end{equation}
For any target-in-camera pose $Z=(R_Z,t_Z)$, define the anchor discrepancy
$r_i(Z)$. Alignment fits the retained anchor/SLAM observations
$\mathcal I_t$ available by update $t$ with uniform confidence:
\begin{equation}
\label{eq:alignmentfit}
\begin{aligned}
r_i(Z)&=\begin{bmatrix}
\lambda\Logv{R_{M_i}R_Z^\top}\\ t_{M_i}-t_Z
\end{bmatrix},\\
\hat A_t&=\argmin_A\sum_{i\in\mathcal I_t}
\bigl\lVert r_i\bigl(S_i(A)\bigr)\bigr\rVert_2^2.
\end{aligned}
\end{equation}
Here $\Logv{R}$ denotes the rotation vector corresponding to the rotation
matrix $R$, with angle in radians.
The factor $\lambda=1\,\mathrm{m/rad}$ balances angular and positional disagreement.
Umeyama alignment~\cite{umeyama1991} of stored translation vectors provides
a provisional initialization, followed by nonlinear refinement with initial
outlier rejection. Later fits start from the previous alignment.

\textbf{Causal fusion.} With alignment fixed, adjacent available SLAM poses
give $\Delta S_{i-1,i}=S_iS_{i-1}^{-1}$. At zero lag, only the current pose
$X_i=(R_{X_i},t_{X_i})$ is free; the previous output $\hat X_{i-1}$ is fixed.
The absolute and relative residuals are
\begin{equation}
\label{eq:fusionresiduals}
\begin{aligned}
e_i^u(X_i)&=g_i\,r_i(X_i),\\
e_i^b(X_i)&=w\begin{bmatrix}
\lambda\Logv{R_{E_i}}\\ t_{E_i}
\end{bmatrix},
\end{aligned}
\end{equation}
where $E_i=\Delta S_{i-1,i}\hat X_{i-1}X_i^{-1}$ has rotation $R_{E_i}$
and translation $t_{E_i}$, and $w>0$ weights relative motion.
The multiplier $g_i=0.05$ when the angle between $R_{M_i}$ and $R_{S_i}$
exceeds $D$, and $g_i=1$ otherwise. This weakens, but retains, a disagreeing
absolute constraint without identifying which branch is wrong.
Missing anchors omit unary residuals; binary residuals require a previous
pose and SLAM at both endpoints. Collecting the available scalar components
in $\mathcal R_i(X_i)$, we solve
\begin{equation}
\label{eq:causalfusion}
\hat X_i = \argmin_{X_i}\sum_{r\in\mathcal R_i(X_i)}h_\delta(r).
\end{equation}
Here $h_\delta$ is the componentwise Huber loss with
$\delta=0.3\,\mathrm{m}$, reducing the influence of outliers; weights multiply
residuals before the loss.

\section{Experimental Evaluation}
We summarize benchmark results before examining SPARK-specific diagnostics.
Section~\ref{sec:ablation} isolates component effects; Sec.~\ref{sec:crossdataset}
discusses results and limitations across datasets.

\subsection{Experimental Setup and Evaluation Metric}
\label{sec:metric}
We evaluate on the SPARK-2024 Stream-2 public
split~\cite{spark2024,rathinam2023spades}, which contains synthetic Unity
renders of PROBA-2 at $T_z=3$--$12\,\mathrm{m}$. MegaPose, PicoPose and DROID-SLAM
use public pretrained checkpoints without target-specific training or
fine-tuning of their network parameters. Temporal parameters are set separately,
as described below. SPARK detection
uses the Mask2Former COCO instance-segmentation release~\cite{cheng2022mask2former}.
The other input conditions are specified in Sec.~\ref{sec:crossdataset}. We report the
SPARK-2024 metric unchanged, consistent with the definitions used by both comparison
methods~\cite{zuo2024crospace6d,sosa2025motion}:
\begin{align}
\Et &= \lVert \hat{T} - T \rVert_2, \quad
\Eq = 2\arccos\!\bigl\lvert\langle \hat{q}, q\rangle\bigr\rvert, \\
\Ep &= \Eq + \frac{\Et}{\lVert T\rVert},
\end{align}
with \Et{} in meters and \Eq{} in radians, computed per frame and averaged within each dataset. \Ep{} is the primary
\emph{pose error} (lower is better), and tables report \Eq{} in degrees. The symmetry-aware
BOP metrics score a $180^\circ$ flip as $\approx\!0$ for a symmetric object and would
obscure the failure studied here. We report the $(\Et,\Eq)$ decomposition alongside \Ep{}
throughout, and use \Ep{} on YCB-Video as well, over its non-symmetric objects
(Sec.~\ref{sec:ycbv}).

\methodmega{} uses MegaPose coarse ranking
and five refinement iterations on $K=10$ candidates. \methodpico{} uses
the three native PicoPose correspondence stages and PnP/RANSAC to produce
$K=5$ candidates, without a MegaPose refiner at inference. Template
poses are converted to the evaluation CAD frame before temporal processing.

Each method uses one temporal configuration across all four datasets.
Table~\ref{tab:parameters} gives the shared benchmark settings for both methods.
Anchors are queried only at available detection opportunities, and
dataset-specific input preparation follows Sec.~\ref{sec:slam}.

\begin{table}[t]
\centering\scriptsize
\caption{Shared temporal configurations for the four-dataset benchmark. Each method uses the same settings across datasets.}
\label{tab:parameters}
\setlength{\tabcolsep}{3pt}
\renewcommand{\arraystretch}{1.15}
\begin{tabular}{@{}>{\raggedright\arraybackslash}p{0.18\columnwidth}>{\raggedright\arraybackslash}p{\dimexpr0.41\columnwidth-2\tabcolsep\relax}>{\raggedright\arraybackslash}p{\dimexpr0.41\columnwidth-2\tabcolsep\relax}@{}}
\toprule
Parameter group & \methodmega{} & \methodpico{} \\
\midrule
Absolute anchors & gap 4, $K=10$, five refiner iterations & gap 6, $K=5$, PicoPose stages 1--3, no MegaPose refiner \\
Decoder & $\alpha=3$, $\mu=0.3$ & $\alpha=100$, $\mu=0.3$ \\
Fusion & $w=0.002$, $D=75^\circ$\newline warm-up 4, Huber scale 0.3 & $w=0.1$, $D=30^\circ$\newline warm-up 4, Huber scale 0.3 \\
\bottomrule
\end{tabular}
\end{table}

The different $\alpha$ values account for the frontends' candidate-score
scales when balancing appearance against motion consistency: MegaPose uses
pose logits, whereas PicoPose uses correspondence/PnP inlier scores.
PicoPose's generally higher pose accuracy and longer anchor gap also motivate
separate tuning of $w$, balancing relative-motion consistency against
anchor-based drift correction.

Both methods retain all temporal stages on every dataset, with gap handling
as specified in Sec.~\ref{sec:fusion}.

\textbf{Alignment updates.} Fusion starts after the anchor/SLAM warm-up in
Table~\ref{tab:parameters}. Alignment is normally refitted every ten fusion
updates using only observations received so far. It is reset when four
successive anchors disagree with the aligned SLAM orientations by more than
$D$, yet agree with each other within $D$ after SLAM rotation propagation.
The new alignment applies from the triggering frame onward; past outputs
remain fixed.

These shared benchmark configurations are distinct from the SPARK-tuned
\methodmega{} configuration used for ablations, range diagnostics and offline
comparisons. Its settings and sensitivity checks are given in Sec.~\ref{sec:hyper}.

\subsection{Pose Accuracy, Error Dispersion, and Throughput}
\label{sec:mainresult}
The shared benchmark \methodmega{} configuration reduces mean pose error
relative to independent MegaPose on all four datasets
(Table~\ref{tab:benchmark-combined}):
$21.3\%$, $6.2\%$, $7.3\%$ and $2.9\%$ on SPARK, YCB-Video, SwissCube
and SHIRT, respectively. \methodpico{} reduces mean pose error relative to
PicoPose on SPARK, SwissCube and YCB-Video, but increases it on SHIRT.

Both versions improve throughput and reduce error SD relative to their
independent estimators on every dataset. SD measures error dispersion, not temporal jitter.
SHIRT's remaining \methodpico{} mean-error regression and SwissCube's candidate limitations are examined
in Sec.~\ref{sec:crossdataset}.

Table~\ref{tab:online} compares causal and offline configurations of both
HAT implementations with the motion-aware ViT on SPARK.

\begin{table}[tb]
\centering\footnotesize
\caption{Causal and offline HAT on synthetic SPARK, with tuned temporal settings.}
\label{tab:online}
\setlength{\tabcolsep}{3.4pt}
\begin{tabular}{@{}>{\raggedright\arraybackslash}p{4.6cm}ccc@{}}
\toprule
Configuration & $\Et\,(\mathrm{m})$ & $\Eq^\circ$ & \Ep \\
\midrule
\textbf{\methodmega{}}, causal & 0.442 & 10.82 & 0.2355 \\
\textbf{\methodmega{}}, all stages offline & 0.430 & 9.66 & 0.2125 \\
\textbf{\methodmega{}}, offline + dense refinement & 0.348 & 9.52 & 0.2016 \\
\midrule
\textbf{\methodpico{}}, causal, tuned & 0.123 & 4.19 & 0.0886 \\
\textbf{\methodpico{}}, offline, tuned & \best{0.104} & \best{2.81} & \best{0.0618} \\
\textit{Motion-aware ViT}~\cite{sosa2025motion}, offline & 0.137 & 4.69 & 0.0990 \\
\bottomrule
\end{tabular}
\end{table}

\subsection{Causal and Offline Performance Comparison}
\label{sec:smoother}
\label{sec:offline}
The causal SPARK-tuned \methodmega{} configuration achieves $\Ep=0.2355$,
versus $0.2125$ offline (Table~\ref{tab:online}): $10.8\%$ higher mean pose
error without future frames or revisions to past outputs. Fresh hypotheses
can correct the current orientation at later keyframes when a suitable
candidate becomes available.

The offline reference uses two global bundle-adjustment passes in
DROID, a decoder backtrace with a whole-sequence gate, and joint
optimization of all frame poses in Stage~4.
Additional dense refinement from the selected anchors reduces the error to
$\Ep=0.2016$ ($\Et=0.348\,\mathrm{m}$, $\Eq=9.52^\circ$). This pass re-acquires
coarse poses when rendered and detected silhouettes diverge and rejects
candidates more than $90^\circ$ from the anchor basin.

Both \methodpico{} variants in Table~\ref{tab:online} achieve lower
translation, rotation and pose errors than the published motion-aware
ViT results~\cite{sosa2025motion}. This offline keypoint method is trained
on the target spacecraft, PROBA-2, using SPADES-RGB and requires a future
frame at inference. \methodpico{} uses fixed pretrained networks in both
its causal and offline variants.

\subsection{Range-Dependent Rotation and Depth Errors}
\label{sec:depth}
\begin{figure}[!hb]
\centering
\includegraphics[width=0.80\columnwidth]{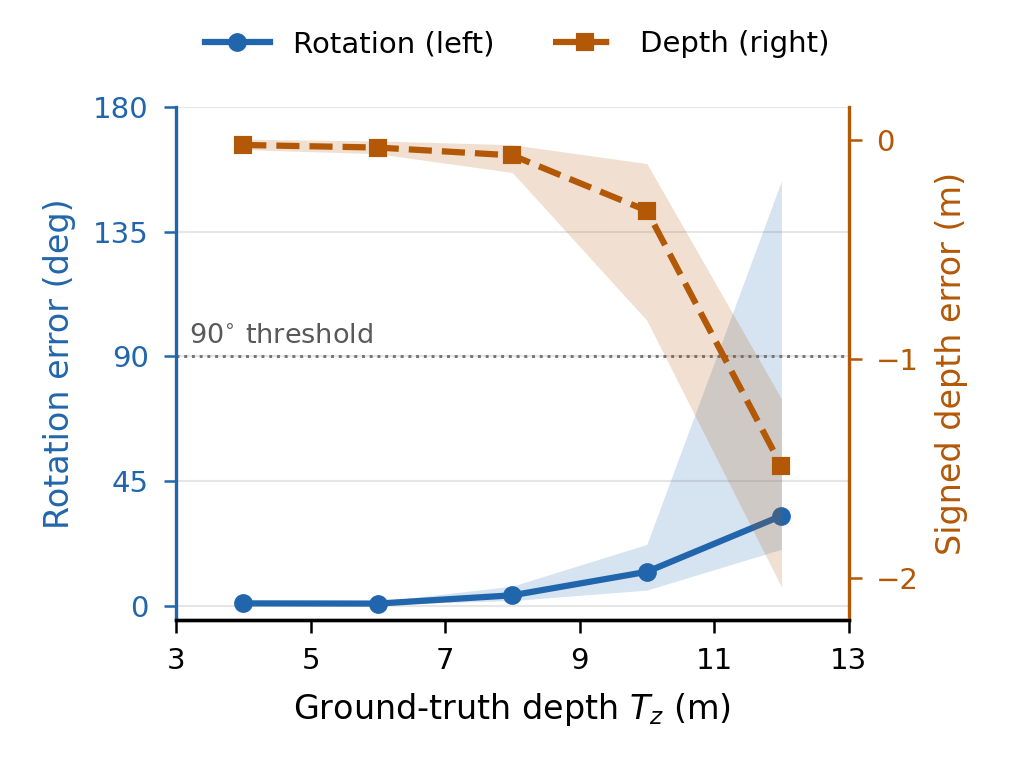}
\caption{Rotation error and signed depth error $\hat{T}_z-T_z$ of
appearance-selected MegaPose hypotheses on SPARK-2024, after refinement
and before temporal selection. Markers show medians in $2\,\mathrm{m}$ bins of
ground-truth depth, and shading shows interquartile ranges.
Negative depth errors indicate underestimated depth.}
\label{fig:range-depth-errors}
\end{figure}

Beyond $\sim7\,\mathrm{m}$, coarse rotation and depth errors increase together
(Fig.~\ref{fig:range-depth-errors}).
Depth contributes $95.5\%$ of translation-error energy, with a $-1.49\,\mathrm{m}$
bias at $11$--$13\,\mathrm{m}$. The $>90^\circ$ coarse-pose share rises from $1.9\%$
below $7\,\mathrm{m}$ to $34.1\%$ at long range. Off-basin hypotheses have median depth
error $-1.34\,\mathrm{m}$ versus $-0.06\,\mathrm{m}$ in-basin. Smaller projections
also erase the details distinguishing near-symmetric orientations. These
observations motivate the size weight in Eq.~\eqref{eq:sizew}.

\section{Ablation Studies}
\label{sec:ablation}
Unless otherwise stated, ablations use the SPARK-tuned configuration of
\methodmega{}, which achieves $\Ep=0.2355$ on SPARK, compared with $0.3254$
for its shared benchmark configuration.
This provides a well-performing reference for assessing each component's
effect on accuracy, in a setting with useful pose candidates and observable
relative motion.

Size weighting reduces appearance influence for small projections.
For box area $a_j$ in pixels squared and apparent size $s_j=\sqrt{a_j}$,
\begin{equation}
\label{eq:sizew}
\begin{aligned}
w_j&=\operatorname{clip}\!\left((s_j/s_0)^\gamma,w_{\min},1\right),\\
\tilde{\ell}_{j,k}&=\alpha w_j\left(\ell_{j,k}-\max_h\ell_{j,h}\right).
\end{aligned}
\end{equation}
Here $s_0=500\,\mathrm{px}$ sets unit weight, $\gamma=2$ its decrease,
and $w_{\min}=0.05$ its floor; $s_j=250\,\mathrm{px}$ gives $w_j=0.25$.
This heuristic favors motion when details occupy fewer pixels; it is not a
calibrated probability. Shared benchmark configurations use $w_j=1$. This selection
weight is distinct from detection-area fusion confidence; Sec.~\ref{sec:hyper}
specifies the loss, warm-up, and fallback settings.

\subsection{Component Contributions}
Table~\ref{tab:evolution} starts with tracking and then adds components to
keyframe-based fusion. The improvement is predominantly rotational.
Basin selection gives the largest proportional reduction between successive
rows, $30.9\%$, followed by gains from size weighting and the disagreement gate. The
size weight is most relevant when the decoder must maintain the selected basin throughout
the far-range segment.

\begin{table}[tb]
\centering\footnotesize
\caption{Cumulative component ablations of \methodmega{} on SPARK, causal throughout.}
\label{tab:evolution}
\setlength{\tabcolsep}{4pt}
\begin{tabular}{@{}>{\raggedright\arraybackslash}p{5.1cm}ccc@{}}
\toprule
Configuration & $\Et\,(\mathrm{m})$ & $\Eq^\circ$ & \Ep \\
\midrule
MegaPose tracking & 0.471 & 28.13 & 0.5387 \\
Keyframe anchors $+$ SLAM fusion & 0.490 & 18.77 & 0.3788 \\
$+$ Viterbi basin selection (Stage~3) & 0.469 & 12.20 & 0.2619 \\
$+$ size-weighted unary & 0.451 & 11.63 & 0.2504 \\
$+$ disagreement gate ($D{=}75^\circ$) & \best{0.442} & \best{10.82} & \best{0.2355} \\
\bottomrule
\end{tabular}
\end{table}

\subsection{Hyperparameter Sensitivity}
\label{sec:hyper}
Unless varied below, settings are stride~10, $K=10$, five refinement
iterations, $\alpha=1$, $\mu=0.3$, $w=0.1$, $D=75^\circ$ and
selection/fusion warm-ups of five/three observations. Size weights follow
Eq.~\eqref{eq:sizew}. The running selection
gate uses a $15^\circ$ neighborhood, minimum median count two and
$70\%$ override limit. Fusion uses detection-area confidence, squared loss
and zero lag. Gaps propagate SLAM rotation and hold translation.

At fixed $K=5$, anchor strides
$5/10/20/40$ give $\Ep=0.2802/0.2843/0.2820/0.3639$.
Strides~5--20 have similar errors at $K=5$, while stride~40 loses accuracy. Increasing
$K$ from 5 to 10 at stride~10 reduces $\Ep$ from $0.2843$ to $0.2355$:
the decoder needs the correct basin at enough keyframes to form a path.

For size weighting, references $s_0=300/400/500/600$ give
$\Ep=0.2399/0.2377/0.2355/0.2371$, versus $0.2402$ without size weighting.
Its independent effect with the fusion gate is small. Without the gate it
reduces $\Ep$ by $0.011$, and at anchors it reduces the $>90^\circ$ share
from $2.83\%$ to $2.40\%$. Gate thresholds $D=45^\circ/75^\circ/120^\circ$
and off give $0.2802/0.2355/0.2358/0.2504$: overly tight gates downweight useful
anchors, while $75^\circ$ and $120^\circ$ perform similarly.

Fusion weights $w=0.05/0.1/0.25/0.5$ give
$\Ep=0.2447/0.2355/0.2868/0.3745$. Non-anchor poses have only relative edges, so
$w$ controls the balance where absolute anchors are present. Excessive weight
propagates SLAM drift, motivating dataset-specific operating points.

\subsection{Continuous-State Filters and Dynamics Priors}
\label{sec:filters}
We replace the discrete decoder with a particle filter on $\mathrm{SO}(3)$, which
can retain competing orientation basins, and a UKF, whose single-Gaussian
representation does not explicitly retain separate modes. Both propagate
with the same SLAM increments, update using the same candidate scores, and
emit a candidate index, leaving the downstream stages unchanged. For each
filter, we report the best-performing configuration found through
hyperparameter search on SPARK.

Over 3,000 keyframes, the discrete forward decoder achieves a mean rotation
error of $11.68^\circ$, with $2.40\%$ of selections above $90^\circ$,
compared with $11.97^\circ$ and $2.93\%$ for the particle filter and
$14.63^\circ$ and $4.73\%$ for the UKF. Even with their best tested
settings, neither filter improves candidate selection on SPARK. The hypotheses form a discrete set regenerated at
each keyframe, and the output must be one of those candidates; the results
therefore favor explicit selection among the available alternatives,
without establishing a general limitation of continuous-state filtering.

A Stage-4 inertial prior penalizing translational second differences and
changes in world-frame incremental rotation leaves $\Ep=0.2355$ unchanged
at the reported precision. These penalties encourage local smoothness,
which is already constrained by the SLAM chain, but do not resolve absolute
orientation ambiguity: a smooth trajectory in a consistently incorrect
basin can remain compatible with the prior. Thus, adding a dynamics prior
does not replace selection among competing orientation hypotheses.

\subsection{Effect of Background Masking on SLAM}
\label{sec:maskablation}
Using raw frames instead of masked targets increases $\Ep$ from $0.2355$
to $0.3481$. Causal DROID fails to recover a trajectory more often
without masking. The textured Earth limb can improve relative rotation on
recovered tracks, but background suppression improves trajectory recovery
across the evaluated sequences.

\begin{table*}[t]
\centering\scriptsize
\caption{Means and sample SD summarize each dataset. Bold denotes the best result in each column per dataset (lowest errors and SD, highest FPS). Italics denote video baselines; our methods appear below the divider. ``+ ref.'' adds five MegaPose refinement iterations. $\mathrm{FPS}$ includes decoding after warm-up and excludes detector generation and loading. Mega-HAT uses the shared temporal configuration. Evaluation and timing protocols are specified in Sec.~\ref{sec:crossdataset}.}
\label{tab:benchmark-combined}
\setlength{\tabcolsep}{1.8pt}
\begin{minipage}[t]{0.495\textwidth}\centering
\textbf{SPARK-2024}\par\smallskip
\begin{tabular}{@{}lrrrrr@{}}
\toprule
Method & $E_p$ & SD & $E_t$ ($\mathrm{m}$) & $E_q$ (${}^\circ$) & $\mathrm{FPS}$ \\
\midrule
MegaPose (per frame) & 0.413 & 0.808 & 0.469 & 21.0 & 0.82 \\
PicoPose & 0.098 & 0.344 & \textbf{0.112} & 4.8 & 2.74 \\
GigaPose + ref. & 0.410 & 0.811 & 0.333 & 21.5 & 1.66 \\
\textit{RGBTrack} & \textit{2.169} & \textit{1.065} & \textit{1.804} & \textit{111.0} & \textit{2.34} \\
\textit{SRT3D + init.} & \textit{0.650} & \textit{0.943} & \textit{1.311} & \textit{29.1} & \textbf{\textit{131.61}} \\
\midrule
Mega-HAT & 0.325 & 0.648 & 0.463 & 15.9 & 2.27 \\
Pico-HAT & \textbf{0.089} & \textbf{0.224} & 0.123 & \textbf{4.2} & 7.03 \\
\bottomrule
\end{tabular}
\end{minipage}\hfill
\begin{minipage}[t]{0.495\textwidth}\centering
\textbf{YCB-Video}\par\smallskip
\begin{tabular}{@{}lrrrrr@{}}
\toprule
Method & $E_p$ & SD & $E_t$ ($\mathrm{m}$) & $E_q$ (${}^\circ$) & $\mathrm{FPS}$ \\
\midrule
MegaPose (per frame) & 0.371 & 0.837 & 0.078 & 16.3 & 0.85 \\
PicoPose & 0.364 & 0.837 & 0.053 & 17.4 & 3.55 \\
GigaPose + ref. & \textbf{0.286} & 0.770 & 0.064 & \textbf{12.4} & 1.87 \\
\textit{RGBTrack} & \textit{1.426} & \textit{1.335} & \textit{0.068} & \textit{77.4} & \textit{5.57} \\
\textit{SRT3D + init.} & \textit{0.722} & \textit{0.938} & \textit{0.083} & \textit{36.2} & \textbf{\textit{94.69}} \\
\midrule
Mega-HAT & 0.348 & 0.761 & 0.083 & 14.7 & 7.07 \\
Pico-HAT & 0.295 & \textbf{0.669} & \textbf{0.051} & 13.6 & 14.81 \\
\bottomrule
\end{tabular}
\end{minipage}\par\medskip
\begin{minipage}[t]{0.495\textwidth}\centering
\textbf{SwissCube}\par\smallskip
\begin{tabular}{@{}lrrrrr@{}}
\toprule
Method & $E_p$ & SD & $E_t$ ($\mathrm{m}$) & $E_q$ (${}^\circ$) & $\mathrm{FPS}$ \\
\midrule
MegaPose (per frame) & 2.113 & 1.027 & 0.160 & 111.1 & 0.76 \\
PicoPose & 1.937 & 5.195 & 0.413 & 70.0 & 3.35 \\
GigaPose + ref. & 2.096 & 1.290 & 0.281 & 101.0 & 1.69 \\
\textit{RGBTrack} & \textit{2.553} & \textit{0.724} & \textit{0.185} & \textit{126.8} & \textit{2.78} \\
\textit{SRT3D + init.} & \textit{2.631} & \textit{0.742} & \textit{0.369} & \textit{126.2} & \textbf{\textit{50.79}} \\
\midrule
Mega-HAT & 1.959 & 0.965 & 0.169 & 101.7 & 2.02 \\
Pico-HAT & \textbf{0.480} & \textbf{0.681} & \textbf{0.057} & \textbf{22.1} & 4.44 \\
\bottomrule
\end{tabular}
\end{minipage}\hfill
\begin{minipage}[t]{0.495\textwidth}\centering
\textbf{SHIRT}\par\smallskip
\begin{tabular}{@{}lrrrrr@{}}
\toprule
Method & $E_p$ & SD & $E_t$ ($\mathrm{m}$) & $E_q$ (${}^\circ$) & $\mathrm{FPS}$ \\
\midrule
MegaPose (per frame) & 1.684 & 1.155 & 1.394 & 83.8 & 0.79 \\
PicoPose & 1.829 & 1.144 & 1.306 & 93.5 & 2.95 \\
GigaPose + ref. & 2.151 & 1.182 & 1.671 & 108.0 & 1.68 \\
\textit{RGBTrack} & \textit{2.733} & \textit{1.259} & \textit{4.122} & \textit{112.7} & \textit{1.68} \\
\textit{SRT3D + init.} & \textit{2.248} & \textit{0.949} & \textit{1.487} & \textit{110.2} & \textbf{\textit{109.79}} \\
\midrule
Mega-HAT & \textbf{1.636} & 1.077 & 1.537 & \textbf{79.8} & 1.06 \\
Pico-HAT & 1.974 & \textbf{0.847} & \textbf{1.058} & 103.8 & 4.82 \\
\bottomrule
\end{tabular}
\end{minipage}\par\medskip
\end{table*}

\begin{figure*}[t]
\centering
\includegraphics[width=0.90\textwidth]{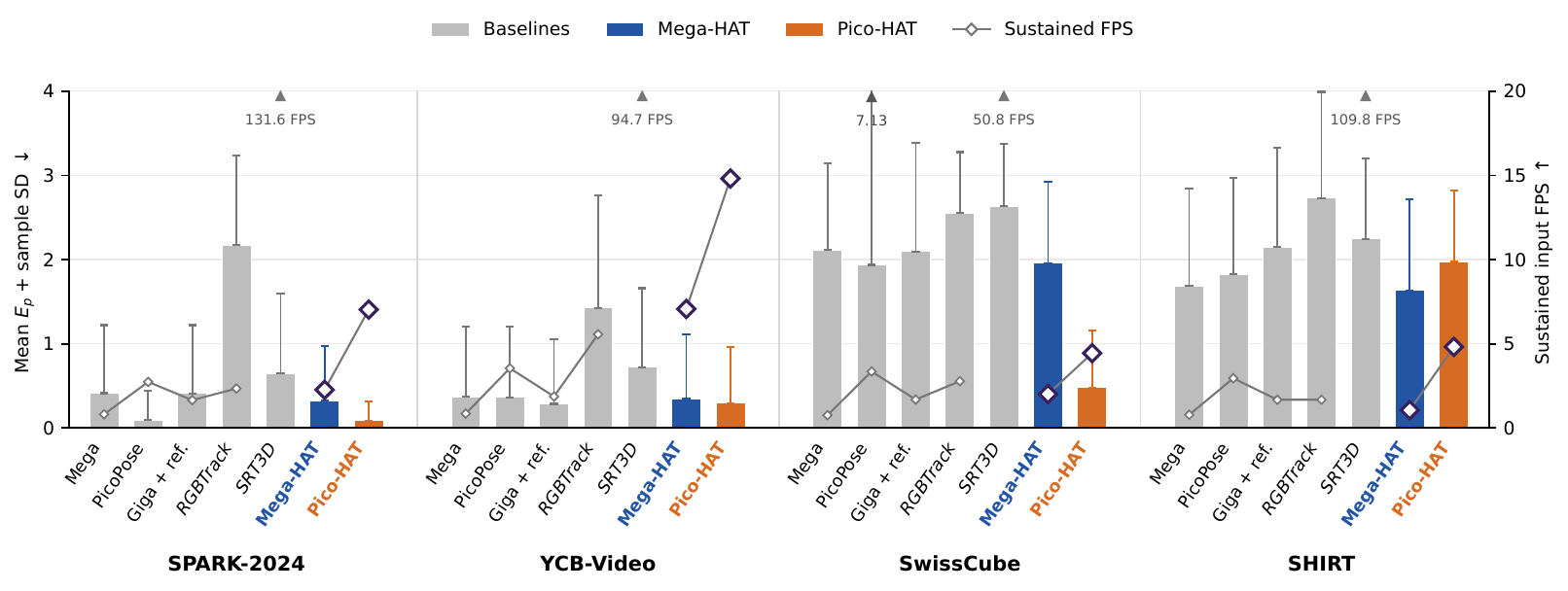}
\caption{Mean $E_p$ (bars), one sample SD (upper whiskers), and sustained input
$\mathrm{FPS}$ (diamonds). Our implementations are colored, with larger outlined diamonds.
Baselines are gray. Numerical labels mark whiskers above 4 and rates above
$20\,\mathrm{FPS}$. Throughput is measured on separate stream subsets.
Table~\ref{tab:benchmark-combined} specifies labels and evaluation scope.}
\label{fig:benchmark-four}
\end{figure*}

\section{Cross-Dataset Evaluation}
\label{sec:crossdataset}
Table~\ref{tab:benchmark-combined} and Fig.~\ref{fig:benchmark-four}
compare both implementations on three spacecraft datasets and YCB-Video.
Video methods process the intervening frames and retain temporal state.
Independent image methods estimate each evaluation image separately. On
YCB-Video, video methods consume the full sequence, while independent methods
run only at the designated BOP target images. SwissCube images come from
its official test split, and SHIRT includes both trajectories in synthetic
and lightbox conditions. We report means and sample SDs for each dataset.

Independent MegaPose restarts coarse
inference at each evaluated image with ten hypotheses and five refinement
iterations. GigaPose supplies five hypotheses to the same refiner, as indicated
by its label. PicoPose selects the highest-scoring valid
native hypothesis without temporal processing or external refinement,
matching the initialization used by \methodpico{}.
SRT3D uses automatic MegaPose initialization. RGBTrack uses RGB-only
initialization and mask-based recovery. All template poses use the verified
CAD-frame conversion. SwissCube and SHIRT use annotation-derived regions
and therefore do not evaluate automatic detection.

Timing uses NVIDIA A100 GPUs ($80\,\mathrm{GB}$) and an Intel Xeon Gold 6330
CPU. Input $\mathrm{FPS}$ counts RGB frames per wall-clock interval after
loading, initialization and warm-up. Models remain resident; detector generation
and offline template preparation are excluded. Each rate is one sustained pass
on a separate stream subset, with fresh pose inference and all temporal stages
executed by both \methodname{} variants.

Both \methodname{} variants exceed their independent frontends' input $\mathrm{FPS}$ on
all four datasets. DROID-SLAM largely determines rate variation through dense
image matching, repeated geometric optimization and per-frame causal pose
readout. Higher input resolution raises SwissCube's cost, while larger image
motion triggers more frequent internal SLAM keyframe updates on SHIRT.

\subsection{Temporal Fusion on SPARK and YCB-Video}
\label{sec:ycbv}
SPARK provides the clearest combination of useful absolute alternatives
and observable relative motion. \methodmega{} lowers both mean error
and SD relative to independent MegaPose, while querying the absolute
estimator sparsely improves throughput. \methodpico{} achieves lower
error and dispersion still, with higher input $\mathrm{FPS}$. This supports using
the temporal framework with different absolute estimators.

YCB-Video~\cite{xiang2018posecnn,hodan2024bop} tests the same temporal
framework on everyday objects outside the spacecraft domain. We evaluate
non-symmetric objects with shared CNOS
detections~\cite{nguyen2023cnos}. \methodmega{} improves mean error
and dispersion relative to independent MegaPose.
\methodpico{} likewise reduces mean error and dispersion relative to independent
PicoPose, while increasing throughput. It also outperforms the shared benchmark
\methodmega{} configuration on these three measures.

\subsection{Candidate Recall Limitations on SwissCube}
\label{sec:swisscube}
SwissCube~\cite{hu2021swisscube} tests a small CubeSat against an Earth
background. A dense MegaPose best-of-ten candidate diagnostic quantifies the remaining
orientation ambiguity: even an oracle selecting the lowest-error hypothesis
has a median rotation error of $65.7^\circ$, and only $22.8\%$ of images
contain a candidate within $30^\circ$ of ground truth. Temporal selection therefore
has limited scope to correct the absolute estimates. The \methodmega{}
configuration improves mean error and dispersion relative to independent
MegaPose. With sparse anchors and uniform confidence, sustained throughput
increases from $0.76$ to $2.02\,\mathrm{FPS}$. Mean pose error remains high at $1.959$,
consistent with the limited orientation accuracy of the absolute hypotheses.

\methodpico{} improves on PicoPose on SwissCube.
Mean pose error falls from $1.937$ to $0.480$, SD from $5.195$ to $0.681$,
and throughput rises from $3.35$ to $4.44\,\mathrm{FPS}$.

\subsection{Rapid-Rotation Challenges on SHIRT}
\label{sec:shirt}
SHIRT~\cite{park2023ukf} exposes a limitation of the DROID-SLAM motion
estimates under rapid spacecraft rotation. The median change in object
orientation between consecutive frames is $4.1^\circ$ on SHIRT,
$1.627^\circ$ on SwissCube, $0.41^\circ$ on SPARK and $0.12^\circ$
on YCB-Video. Larger inter-frame rotations challenge correspondence estimation
and can cause DROID to underestimate motion, weakening the temporal
constraints used for basin selection and fusion.

As summarized in Table~\ref{tab:benchmark-combined}, \methodmega{} provides
only a modest accuracy gain on SHIRT, while \methodpico{} increases mean pose
error despite reducing dispersion. Smoother estimates can therefore remain
in an incorrect orientation basin when relative motion is poorly observed,
highlighting the need for more reliable motion estimation under rapid tumbling.

\section{Conclusion}
This paper presents \methodname{}, a method for monocular spacecraft pose estimation
that uses relative motion to select among competing absolute-pose hypotheses
before alignment and fusion, without revising past outputs. Its two versions
use MegaPose or PicoPose with a metric CAD model and require no target-specific
training or fine-tuning of the pretrained networks. Using one temporal
configuration per version, both increase sustained input $\mathrm{FPS}$ and
reduce error dispersion relative to their independent estimators on all four
datasets. Both also reduce mean pose error, with the sole exception of
\methodpico{} on SHIRT. The ablations identify basin selection before fusion
as a principal source of the accuracy gain.

\textbf{Limitations and future work.} Absolute orientation discrimination
degrades at long range as target details occupy fewer pixels, while rapid
rotation can degrade SLAM and invalidate the relative-motion constraints,
as SHIRT demonstrates. Improved basin discrimination for small target
projections and relative-motion estimation robust to rapid tumbling are
therefore priorities for future work. The framework requires a metric
CAD model, useful absolute-pose candidates and inter-frame motion that
SLAM can observe. SwissCube highlights the candidate-recall limitation:
temporal selection cannot recover an orientation absent from the candidate
set.

\setlength{\IEEEilabelindent}{0pt}

\FloatBarrier

\end{document}